# A COMPUTER VISION APPROACH TO ESTIMATE THE LOCALIZED SEA STATE

Aleksandar Vorkapic, Miran Pobar, Marina Ivasic-Kos

*Faculty of Informatics and Digital Technologies, & Centre for Artificial Intelligence, University of Rijeka, 51000 Rijeka, Croatia*
a.vorkapic@icloud.com; mpobar@uniri.hr; marinai@uniri.hr

## Abstract

This research presents a novel application of computer vision (CV) and deep learning methods for real-time sea state recognition, aiming to contribute to improving the operational safety and energy efficiency of seagoing vessels, key factors in meeting the legislative carbon reduction targets. Our work focuses on utilizing sea images in operational envelopes captured by a single stationary camera mounted on the ship bridge. The collected images are used to train a deep learning model to automatically recognize the state of the sea based on the Beaufort scale. To recognize the sea state, we used 4 state-of-the-art deep neural networks with different characteristics that proved useful in various computer vision tasks: Resnet-101, NASNet, MobileNet_v2, and Transformer ViT -b32. Furthermore, we have defined a unique large-scale dataset, collected over a broad range of sea conditions from an ocean-going vessel prepared for machine learning. We used the transfer learning approach to fine-tune the models on our dataset. The obtained results demonstrate the potential for this approach to complement traditional methods, particularly where in-situ measurements are unfeasible or interpolated weather buoy data is insufficiently accurate.

This study sets the groundwork for further development of sea state classification models to address recognized gaps in maritime research and enable safer and more efficient maritime operations.

## 1. Introduction

Accurate assessment and anticipation of sea states are important for ensuring navigational safety and minimizing a ship's energy consumption. The planning of safer and more fuel-efficient routes is facilitated by the prediction of sea conditions, which allows for the avoidance of areas with adverse weather that could lead to increased Greenhouse Gas (GHG) emissions from excessive carbon-based fuel consumption. The International Maritime Organization (IMO) leads such initiatives in the maritime sector, targeting a reduction of GHG emissions and fostering global climate change mitigation. Specifically, the IMO's GHG emissions strategy sets a target to decrease the carbon intensity of all ships by 40% by 2030, using 2008 as a benchmark. The short-term measures promoted to reach these targets include adopting more energy-efficient vessels to reduce GHG outputs. Fuel consumption for propulsion constitutes the largest portion of a vessel's total energy distribution; therefore, optimizing propulsion leads to the most significant energy savings (Lindstad, et al. 2015) (Bouman, et al. 2017) (Waleed M. Talaat 2023). Effective navigation routes mapped out with consideration for adverse weather can further mitigate the impact on a ship's performance (Wu, et al. 2021). By integrating weather and oceanographic data into their strategic planning, ship operators can enhance navigational safety, optimize operational costs, and lessen the environmental footprint, advancing towards the IMO's set energy efficiency and sustainability in maritime operations (Ruth and Thompson 2022). Operational parameter predictions, which are essential for efficient route planning, currently utilize machine learning systems whose accuracy is contingent upon the quality of input data. Reliance on meteorological information, though valuable, may lack the precision necessary for optimal predictions. Moreover, wind speed sensors, which provide critical data inputs, are often not installed on smaller vessels. The integration of localized sea state data into these machine learning models enriches the dataset, leading to a substantial improvement in the model's learning capability and prediction accuracy. Consequently, this refined approach enables more informed route planning, helping to reduce the environmental impact of shipping and supporting the maritime industry's progress towards achieving its energy efficiency and sustainability targets.

Autonomous maritime operations are gaining interest in maritime-related research, offering the potential to lessen human error and enhance both safety and efficiency. Computer vision (CV) is emerging as a key technology for autonomous shipping, enabling ships to detect and respond to their environment in real time (Moosbauer, et al. 2019) (Qiao, et al. 2021). By employing CV-based sea state recognition, ships can make informed decisions and adjust their operations, leading to safer and more efficient operations. Additionally, CV also brings promise to the fields of weather and oceanographic forecasting. The ability to recognize sea states in real time can contribute to more accurate, timely forecasts, enabling the marine industry to plan navigation more effectively and safely.

There has been interest in the application of machine learning techniques to determine sea states, aiming to achieve more accurate and reliable results. An innovative concept for shipboard sea state estimation through the measurement of wave height, wave period, and wind speed is presented by (Nielsen, et al. 2015). The study provides insights into the use of sensor data for sea state estimation and highlights the significance of accurate sea state identification for maritime safety. A sea state classification method based on ship movement four-degree-of-freedom (DoF) ship motion data (surge, sway, roll, and yaw) was developed by (Tu, et al. 2018). The data were pre-processed and decomposed into 20 categories per DoF, and 11 statistical features were extracted to train their model. Their classification system was built in three layers: the first used adaptive neuro-fuzzy interference system classifiers for each category, the second involved random forest classifiers, and the third applied a bias-compensating method for averaging the outputs with reported accuracy ranging from 74.4% to 96.5%. Notably, this method is dependent on ship motion sensor data and did not extend to the sea state classification through visual sea images. Application of multifractal methods for the analysis of sea surface images for sea state determination is introduced by (Ampilova, et al. 2019). The study provides insights into the use of fractal-based approaches for sea state identification and highlights the importance of considering the multiscale nature of sea surface patterns. A novel time-frequency image-based approach for sea state estimation using motion data from dynamically positioned vessels is proposed by (Cheng, et al. 2019). The study utilized simulated ship movement data from nine sensors to train and test a convolutional neural network (CNN) and highlighted the potential of data-driven approaches for sea state estimation. By employing a short-time Fourier transform, each sequence of data from the sensors was converted into a single spectrogram. The sea states were bifurcated into 5 categories, and the proposed 2D CNN model was trained to classify them using the spectrogram. The results were compared to long-short-term memory architecture (LSTM) and 1D CNN models, with the proposed model achieving the highest classification accuracy of 94%. However, the study did not report the number of spectrogram instances per class and in the overall dataset, as well as the per-class classification accuracy. Consequently, the model's generalization ability across various classes remains unknown. A sea state classification method is proposed by (Liu, et al. 2021) using a LeNet-based model and sea clutter radar data from two sources. The datasets covered three sea states with significant wave heights ranging from 0.5 m to 4 m. Separate models were trained and tested on the datasets, resulting in classification accuracies of 95.75% for states 3–4 and 93.96% for states 4–5. The lower accuracy (93.96%) was attributed to limited buoy measurement calibration points for the corresponding sea clutter radar data. Although the study claims strong generalization ability, the models were tested in a small temporal and geographical setting, and high-class imbalance was observed in the datasets. Consequently, the models may exhibit bias towards the majority class. Residual network, ResNet-152 model to classify sea states from optical images of the sea surface is developed by (Zhang, Yu and Qu 2021). The study divided sea states into 10 categories based on ship movement and waveforms, using video data collected from a ship-mounted visual-range video camera. An unusual data split ratio of 82:18 was employed for training and validating the model, which achieved a validation accuracy of 89.3%. However, the study does not report any measure to quantify the testing accuracy of the proposed method, nor does it mention sea-state-wise and overall classification accuracy. Additionally, the method's division of sea states into 10 categories does not follow recognized standards such as the Beaufort or Douglas scales. Consequently, the study has limited significance. A recent study by (Umair, et al. 2022) introduced a deep learning model for sea state classification using visual-range sea images in four sea states ranging from 1 to 4 on the Beaufort scale. The model demonstrated a high accuracy rate of 97.8% on a test dataset, showing promising results. The study highlights the potential of deep learning in sea state recognition and the relevance of visual-range imagery in such applications. Another recent study was conducted by (Mittendorf, et al. 2022), who used in-service data from a container vessel to compare the performance of various machine learning algorithms for sea state identification. They found that a random forest algorithm achieved the best results, with an accuracy of 90.5% on a test dataset. The study provides valuable insights into the use of machine learning for sea state identification based on in-service data. Development of the HIDRA2 model for sea level and storm tide forecasting in the northern Adriatic, using a deep-learning ensemble trained on a large dataset of sea level and meteorological data is proposed by (Rus, et al. n.d.). HIDRA2 outperformed other state-of-the-art models, achieving a root mean square error of 0.089 m for sea level and 0.111 m for storm tides. Including such information significantly improved the model's performance. HIDRA2 represents an important advance in sea level and storm tide forecasting, with implications for coastal management and disaster preparedness in the region.

The relevant research has revealed challenges in accurately identifying sea states, particularly in the diverse and dynamic conditions encountered in real-environment applications. A notable gap is the lack of comprehensive datasets that reflect the navigating range of sea states. These present opportunities for further research to harness CV techniques and develop algorithms that can effectively interpret sea imagery within the operational envelope.

The specific objective of this research is to address the aforementioned gaps by creating and developing a CV algorithm capable of accurately detecting and categorizing various sea states in real operational conditions. This approach employs cameras and deep learning algorithms to automatically classify waves, providing a standardized measure of sea state. It can be particularly valuable in areas where traditional methods, such as in-situ measurements, are not feasible or cost-effective (Vorkapić, Radonja and Martinčić-Ipšić 2021), or where interpolated values from weather buoys are insufficiently accurate. The CV-based sea state classification approach could potentially be used alongside physical methods to provide a more comprehensive understanding of seagoing dynamics.

The main contributions of our work can be summarized as follows:

    1. Original custom dataset with 8 sea state classes according to Bf 1 to 8 prepared for supervised machine learning,

    2. Fine-tuned ResNet, MobileNet, ViT, and NASNet deep models for sea state recognition on our original data set,

    3. Proposal of image sampling strategies in the case of texture recognition or small object detection and analysis of the results according to the sampling strategy,

4. Comparison of model performance according to standard metrics and selection of the most suitable models for the task of recognizing the state of the sea.

The paper is structured as follows: Section 1. presents the problem, and research objectives and provides a review of the previous research; Section 2. presents the background of the dataset, tools, evaluation methods, and modeling process; Section 3. validates machine learning methods and illustrates the results obtained, followed by Section 4., which concludes the paper.

## 2. Material and methods

The Beaufort scale is a traditional method for estimating sea state, linking wind speed to observed conditions at sea or on land. It categorizes the sea states into 13 classes (0-12) based on the corresponding wind speeds and provides a brief description of the visual features of each state. The process is typically carried out through visual observation from the bridge, but the potential for human error makes computerized sea state classification models an appealing alternative. Utilizing CV for sea state recognition based on the Beaufort scale offers an efficient method for monitoring sea state and wave conditions.

### 2.1. Dataset collection

To train the models for sea state recognition, as part of this research study, an extensive set of data that includes the target sea states had to be first collected and then annotated and prepared for machine learning. The collected raw data encompasses 119 video recordings, which collectively constitute 39.61 gigabytes of high-definition visual data. These recordings were made in 11 distinct geographic locations, ensuring a broad coverage of maritime environments. Utilizing a stationary camera with a resolution of 3840x2160, the footage was captured without human intervention apart from starting and stopping the recordings and saved using the HEVC codec. The positioning of the camera on the left bridge wing of the vessel afforded a panoramic view of the sea and included segments of the ship, the wake, and the distant horizon. The height at which the recordings were captured varied with the vessel's loading condition: the bridge wing stood at 38.12 meters above sea level while in cargo, and 40.32 meters when in ballast.

Given the inherent challenges in documenting sea states above 6-7, the vessel of choice for this study was a large ocean-going ship, specifically selected for its operational capabilities in the harsh winter conditions and rough seas. This selection was aimed at maximizing the probability of capturing recordings of the higher sea states. The video recordings were conducted on an 89,432 DWT and 173,998 m3 liquefied natural gas transport vessel, akin to the vessels from a 174 km3 series previously constructed by Korean shipbuilders. The recording period spanned from 12/02/2022 to 24/12/2022, during which the vessel was both in navigation and at anchor, to ensure a diverse representation of sea states could be captured and analyzed.

The exact geographic locations of the recordings are shown in Figure 1.

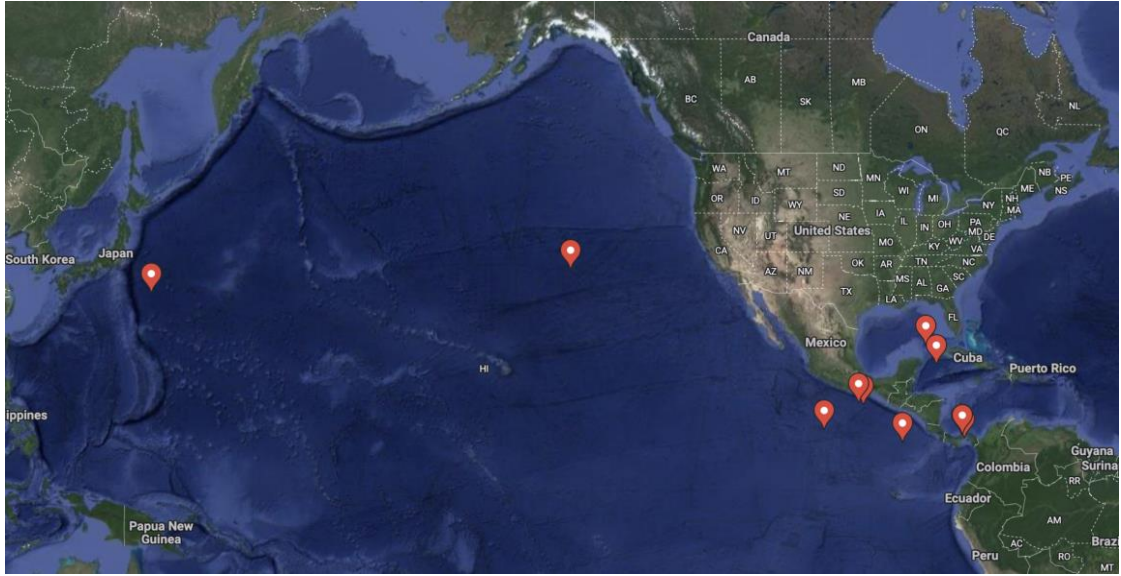

**Figure 1. Map of sea state image acquisition locations.**

Figure 2 presents two representative frames from the video footage, illustrating the type of raw visual data obtained from the stationary camera mounted on the vessel's port bridge wing. These frames show the specific maritime environment section that was consistently in view, demonstrating how the camera's positioning captures a standardized area across various recordings. The selected individual crops focus on key sea state features excluding areas that offer less analytical value, such as most of the sky or the ship's wake. The decision to mount the camera on the port side, following initial testing including trials at the bow, was made for its optimal accessibility, the visibility it provides of the ship's structure with known dimensions, and the side perspective it offers during maneuvering in narrowed or congested waters.

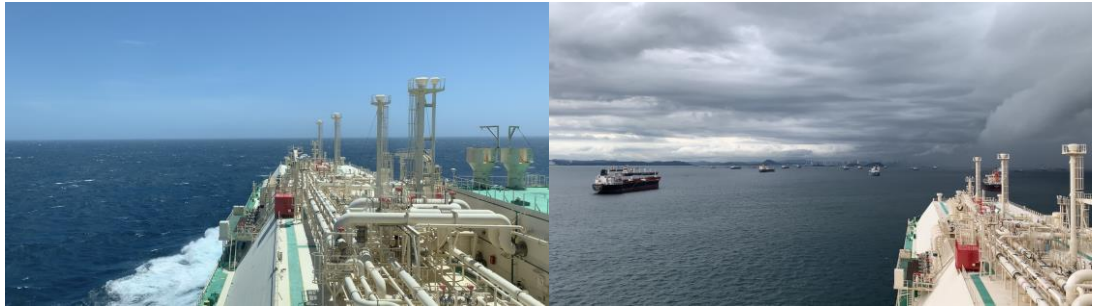

**Figure 2. Sample frames for maritime video dataset for sea state analysis.**

For each recorded video, the sea state was estimated on the Beaufort scale by human experts as would occur during regular ship operations. These estimations, while mostly corresponding with the ship's anemometer wind speed measured at a height of 50 meters, are subject to variability due to the multifaceted nature of wave formation, including wind duration, gusts, and existing swell. Thus, anemometer readings are not completely consistent with sea state conditions and may not always provide a reliable basis for these estimations. To address this, our computer vision models are trained on these expert estimations as target values, aiming to utilize a large and diverse dataset to refine the accuracy and consistency of sea state assessments, reduce the variability inherent in human estimations, and assist and potentially improve the precision of human operators.

The basic data about the collected videos for each sea state is shown in Table 1.

**Table 1. Statistics of collected videos for sea states from 1-8 Bft.**

| Sea state | 1 | 2 | 3 | 4 | 5 | 6 | 7 | 8 |
|---|---|---|---|---|---|---|---|---|
| No. sessions | 6 | 7 | 7 | 7 | 7 | 4 | 3 | 3 |
| No. frames | 19.173 | 24.101 | 14.030 | 33.063 | 46.303 | 16.282 | 15.094 | 6.393 |
| Total duration (m:s) | 5:20 | 6:42 | 3:54 | 9:11 | 12:52 | 4:31 | 4:12 | 1:47 |

Within the scope of our dataset, we were successful in capturing sea states spanning from 1 to 8 on the Beaufort scale. The total duration quoted in Table 1 represents the periods of interest for our study, focusing on capturing a diverse range of these sea states. The absence of state 0 and those beyond 8 in our dataset is attributable to operational challenges in obtaining such data; state 0, or 'calm' conditions, are relatively rare in the operating environment of the oceangoing vessel, and states above 8 are exceedingly difficult to capture due to their severity and the associated safety concerns. Consequently, our dataset provides a significant range of conditions for analysis given that, as a rule, ships navigate in these conditions. Sea state 8 is already considered borderline for operational safety and energy efficiency on the subject vessel, and sea states above 8 are actively avoided. This targeted approach to data collection ensures that the computer vision model is trained on varied conditions that are most reflective of operational scenarios, which explains the shorter durations recorded compared to the full 10-month period. Given the usual requirements for a robust data set, we acknowledge that the samples presented in Table 1 include relatively short video segments due to our selective recording process focused on collecting different relevant sea states in regular maritime operations and trying to form a balanced data set that is then actually limited to the samples collected for the rarest class, class 8.

### 2.2. Dataset preparation

The basic prerequisite for the implementation of this research is the definition of a customized dataset intended for the development and validation of machine learning models aimed at sea state analysis that is formed from the collection of video recordings. From the collected raw video material, a customized data set for training and testing the model was constructed.

The initial collection of videos was found to be imbalanced concerning the total number of frames, with certain classes containing more data than others. To form the balanced dataset from the initial uneven quantity of data for each sea state, and to distill the video data into a form more suitable for analysis, a systematic sampling approach was employed. This process involved the selection of frames at regular intervals from the recorded footage. A sampling interval was chosen for each class individually so that the resulting number of frames for each class is approximately the same. From each chosen frame, a segment with dimensions of 331x331 pixels was extracted from the bottom-left corner of the frame (Fig. 3). This size was chosen as it is the native size for the NASNet network and larger than the default input sizes of other used networks in this work (commonly 224x224). Also, larger crops from the source frames would greatly increase the presence of the ship's wake in the image, which already appeared sometimes with 331x331 crops, so segments had to be checked manually and such crops removed. The extracted segments were then incorporated into the training subset with about 750 images per class (Table 2). This number was chosen after a preliminary test with different training set sizes showed that the performance of the model increased with additional images but didn't improve further after about 300 images per class, as described in Subsection 3.2. Test and validation subsets were created similarly, containing about 300 images per class that were not included in the training set. This dataset was named UNIRI-SeaState-LL. An illustrative set of samples representing each sea state in the resulting dataset is shown in Figure 5.

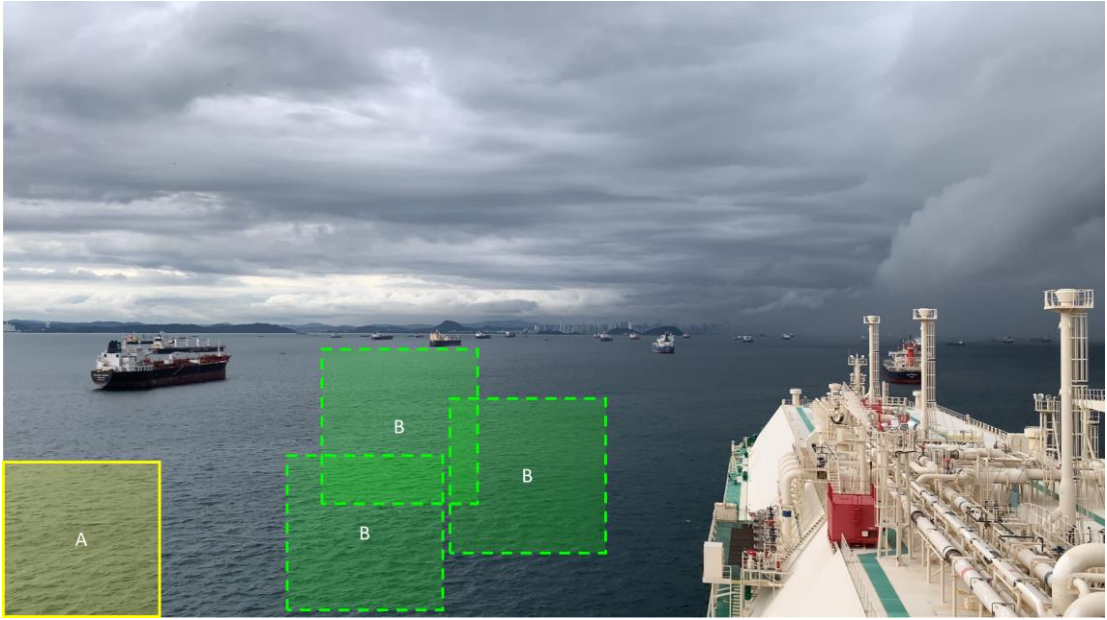

Figure 3. The size and location of the cropped portion of the initial frame are included in the dataset UNIRI-SeaState-LL (Yellow frame marked A). Green frames marked B represent some possible locations of crops included in the dataset UNIRI-SeaState-R.

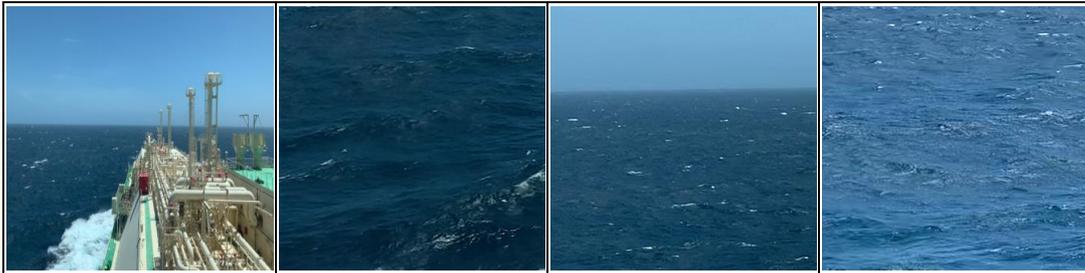

Figure 4. Example of sea state 7 in a) resized image b) crop from the bottom-left part of the image, c) crop near the horizon line, d) crop from a random part of visible sea.

The rationale behind the frame size and sampling frequency is directly correlated with the model input size and the height from which the sea was recorded. Resizing images to fit the input dimensions of our models resulted in the loss of details needed for accurate sea state determination (Fig. 4a). Furthermore, the recordings were obtained from heights of 38.12 meters and 40.32 meters, depending on the vessel's loading condition. At such elevations, the discernible differences between adjacent sea states are marginal and can be challenging to detect. To overcome this, we explored various image sampling strategies that would allow us to preserve important features within the constraints of model input sizes, namely cropping from the bottom-left corner (Fig. 4b), cropping near the horizon (Fig. 4c) and random crops within the part of the image with the sea visible (Fig. 4d). This zoomed perspective is instrumental in enhancing the visibility of features that are critical to the accurate classification of sea states, thereby facilitating a more nuanced analysis conducive to the objectives of this study. The crops near the horizon line contained less visible detail and were difficult to obtain reliably with a significant proportion of sea to the sky, so were not included in the further experiments.

In the development of the dataset, we have also explored a variant where the sample is taken from a random position to the left of the ship's hull (Fig. 3). This variant of the dataset is named UNIRI-SeaState-R.

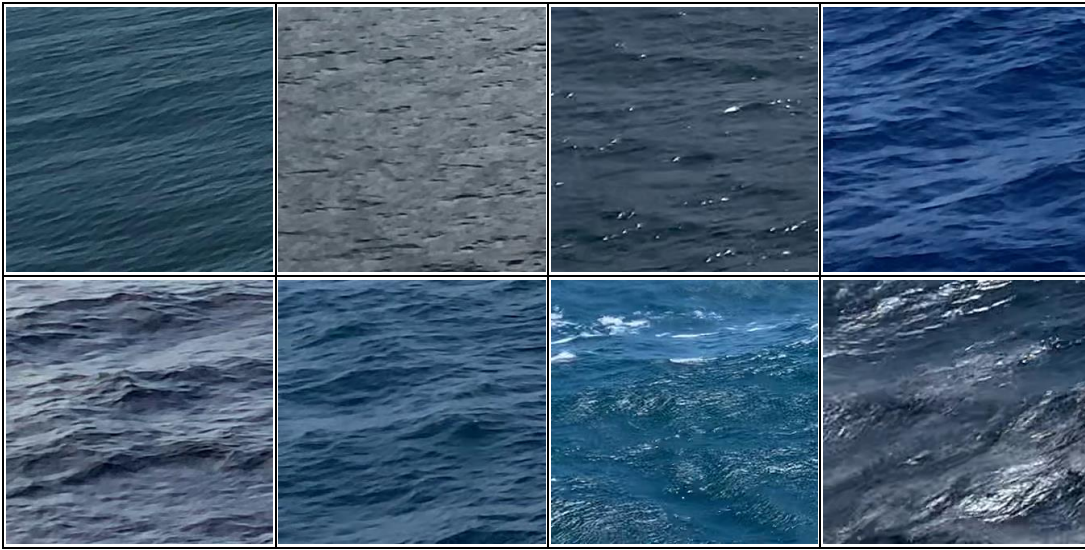

**Figure 5. Examples of extracted sea state images. Top row: states from 1 to 4 Bft. Bottom row: states from 5 to 8 Bft.**

**Table 2. Class Distribution Balance Across Different Cropped Image Datasets**

| Dataset | Classes | | | | | | | | Total images |
|---|---|---|---|---|---|---|---|---|---|
| | 1 | 2 | 3 | 4 | 5 | 6 | 7 | 8 | |
| UNIRI-SeaState-R | 750 | 750 | 737 | 750 | 750 | 744 | 710 | 766 | **5.957** |
| UNIRI-SeaState-LL | 750 | 736 | 750 | 750 | 750 | 758 | 706 | 770 | **5.970** |

Table 2 details the resultant class distribution for the training subset of both datasets. It provides a numerical overview of how images were equitably distributed across the classes to facilitate a balanced representation in line with the Beaufort scale's defined sea states.

## 2.3. Model architectures

In the development of models for sea state classification, our selection of model architectures was guided by the proven efficacy of contemporary classifiers in the realm of machine learning. We chose to integrate renowned architectures such as ResNet-101 (He, et al. 2016), NASNet (Zoph, et al. 2018), MobileNet V2 (Sandler, et al. 2018), and Vision Transformer (Kolesnikov, et al. 2021), all of which have been trained on extensive datasets like COCO (Lin, et al. 2014) or ImageNet (Deng, et al. 2009). These models have not only excelled in benchmark tests but have also set the standard for accuracy in diverse applications.

We adopted the architecture of these classifiers and used fine-tuning on our dataset from the pre-trained weights to build models that can reliably interpret the complex and dynamic nature of maritime environments.

The used model architectures are described in brief in the following subsections.

**ResNet-101**

ResNet-101 is a deep convolutional neural network (CNN) architecture designed to address the vanishing gradient problem in very deep networks. It belongs to the ResNet (Residual Network) family, which was introduced by Microsoft Research in 2015. The "101" in its name refers to the number of layers in the network. The main innovation in ResNet is the introduction of residual blocks, which include skip connections or shortcuts to facilitate the flow of gradients during training. This helps mitigate the vanishing gradient problem and allows for the training of significantly deeper networks.

ResNet-101 has proven to be highly effective for various computer vision tasks, including image classification, object detection, and segmentation. Its architecture has become a foundation for the development of even deeper and more complex neural networks, contributing to advancements in the field of deep learning.

The basic building block of ResNet-101 is called the bottleneck residual block (Fig. 5). The block comprises three layers, a 1x1 convolution, 3x3 convolution, and another 1x1 convolution layer, where the first 1×1 layer reduces, and the final 1x1 layer restores the dimensions, leaving the middle 3×3 layer with smaller input/output dimensions, effectively reducing the computational cost.

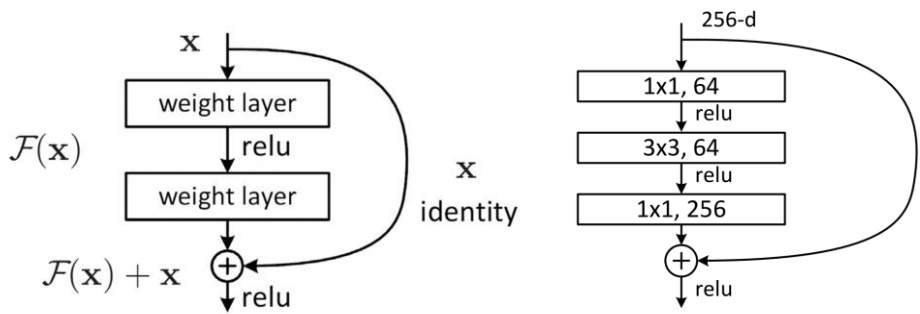

**Figure 6. A "bottleneck" building block for ResNet-50/101/152 (He et al. 2016)**

The key element of the residual block is the shortcut connection, or residual connection, which adds the input of the block directly to the output. This shortcut allows the gradient to flow directly through the network.

ResNet-101 has a total of 33 such building blocks, and with an additional 2 layers at the input, contains a total of 101 layers.

At the end of the network, ResNet-101 employs a global average pooling layer followed by a fully connected layer with a softmax activation function for classification tasks.

## NASNet

Similar to ResNet, NASNet architecture is formed by stacking a fixed structure of layers that form basic building blocks to form the complete network. However, the structure of the building blocks is not designed manually but discovered automatically using a reinforcement learning-based process, called neural architecture search (NAS). The neural architecture search process employs a reinforcement learning agent, guiding the selection of operations within the building blocks (cells) based on their performance in specific tasks.

The building blocks of NASNet-A were determined on the classification task using the CIFAR-10 dataset and employed successfully for classification on other datasets like ImageNet.

The architecture includes two types of cells: reduction cells and normal cells (Fig. 6.). Reduction cells downsample spatial dimensions, reducing the spatial resolution, and normal cells maintain the spatial resolution.

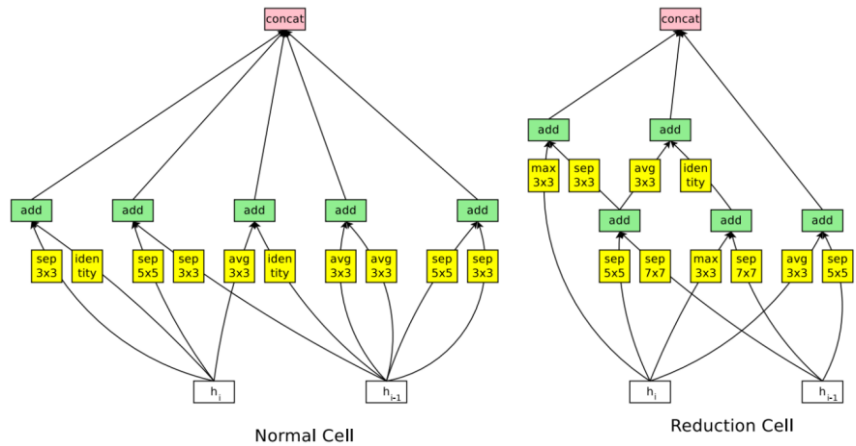

**Figure 7. Normal and Reduction cells of NASNet. (Zoph et al. 2018)**

NASNet-A focuses on efficiency and is designed to achieve state-of-the-art performance with fewer parameters and computations compared to manually designed networks.

The neural architecture search process considers multiple objectives simultaneously, such as accuracy, model size, and computational efficiency. This multi-objective optimization results in a balanced trade-off between performance and resource requirements.

NASNet-A has demonstrated competitive performance on benchmark datasets, achieving top-tier accuracy in image classification tasks while being computationally efficient.

Based on the NASNet-A normal and reduction cells, many configurations of the network with different numbers of stacked cells and different numbers of filters in the initial convolutional cell have been used, offering trade-offs between computational complexity and performance.

In this work, a configuration with four repetitions of normal cells between reduction cells and 1056 filters in the initial layers was used. This is a lightweight configuration with only about 5 million parameters, suited for mobile and edge computing devices.

The architecture aims to strike a balance between achieving satisfactory performance and minimizing the computational resources needed for inference on mobile platforms.

### MobileNet V2

MobileNetV2 is a lightweight convolutional neural network architecture designed for environments with constrained resources such as mobile and edge devices. It is an evolution of the original MobileNetV1, focusing on design simplicity and improved performance.

MobileNetV2 introduces the new inverted residual layer block that uses a lightweight depthwise separable convolution followed by linear bottlenecks (Fig. 7). The depthwise separable convolution is computationally much simpler than the full convolution while retaining good performance. The bottleneck layer is a 1x1 convolution with a linear activation function that helps preserve information flow. Similar to the building block of ResNet, the inverted residual layer of MobileNetV2 features skip connections, however here the skip connections are between the input and the lower-dimensional bottlenecks. In each inverted residual block of MobileNetV2, an expansion layer within the depthwise separable convolution allows for an increase in the number of channels, controlled by the expansion factor t, enabling the network to capture more complex features. The linear bottleneck layer following the depthwise separable convolution reduces the dimensionality back to facilitate information flow through skip connections.

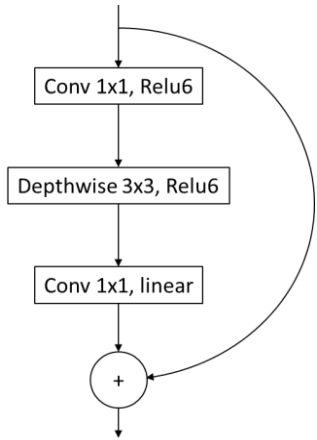

**Figure 8. Inverted residual block of MobileNetV2, adopted from (Sandler et al. 2018)**

Collectively, these features contribute to its suitability for deployment on mobile and edge devices, enabling inference with reduced computational demands.

**Vision Transformer**

The Vision Transformer (ViT) model is a departure from the conventional convolutional neural network-based models. ViT leverages the transformer architecture, originally designed for natural language processing (NLP), and applies it directly to image data. As the standard Transformer receives a 1D sequence of token embeddings, the 2D images are first reshaped to conform to the expected shape. The input images are divided into fixed-size non-overlapping patches and flattened to 1D. Subsequently, the sequences of flattened pixel values are treated as "tokens" in NLP applications. The patches are linearly embedded into high-dimensional vectors, forming the input sequence for the transformer.

ViT uses a standard transformer architecture, consisting of self-attention mechanisms and feedforward neural networks. The self-attention mechanism enables the model to capture global dependencies within the input sequence. Since transformers do not inherently understand the spatial layout of the input image, the positional information in the form of positional embeddings is explicitly added to the patch embeddings to preserve the spatial information of the image.

Furthermore, the lack of translation equivariance and locality, which is a feature of CNNs, hinders the capability of the transformer to generalize well when trained on insufficient amounts of data. Smaller datasets can be used successfully using fine-tuning from models that are pre-trained on larger datasets.
The ViT architecture is scalable, and in this work, the ViT-b32 variant is used. This is a smaller model with about 87 million parameters, in comparison with the ViT-Large variant with more than 300 million parameters. The number of parameters is still much larger than other networks tested, especially the NASNet Mobile and MobileNetV2 architectures.

## 2.4.  Experiment setup

The four model architectures described above were each trained on the two variants of the dataset, UNIRI-SeaState-LL, and UNIRI-SeaState-R, resulting in eight total trained models. All models were trained in two steps from initial weights pretrained on the Imagenet dataset. On top of the base models, a global average pooling layer and a dense layer with linear activation were used, except for the ViT-b32 model, where a dense layer with GELU activation function was used before the final prediction layer with linear activation.

First, the models were trained using transfer learning for 30 epochs, starting from the ImageNet pretrained weights and the Adam optimizer was used with a fixed learning rate of 0.0001 and categorical cross-entropy loss function. In this stage, the weights in all layers of the network were frozen except for the final (output) dense layer. Then, in the second step, several layers were unfrozen to allow fine-tuning of a part of the base network on new data, as indicated in Table 3. The resulting number of trainable parameters in this step is also shown in Table 3 for each architecture. For the second step, the RMSProp optimizer was used with categorical cross-entropy loss. The number of training epochs varied from 200 to 1000 between different architectures, chosen according to the point where there was no further improvement in the training accuracy. In the second step, a progressive learning rate was used, with a base learning rate of 0.0001 and with a reduction on a plateau by a factor of 5 down to a minimum learning rate of 1e-6, with patience of 30 epochs and a threshold of validation accuracy improvement of 1e-6.

The batch size for all models was 250, except for ViT, where 200 images were used in a batch due to memory constraints. As an example, training and validation accuracy for ResNet trained on the UNIRI-SeaState-random-331 dataset is shown in Figure 8.

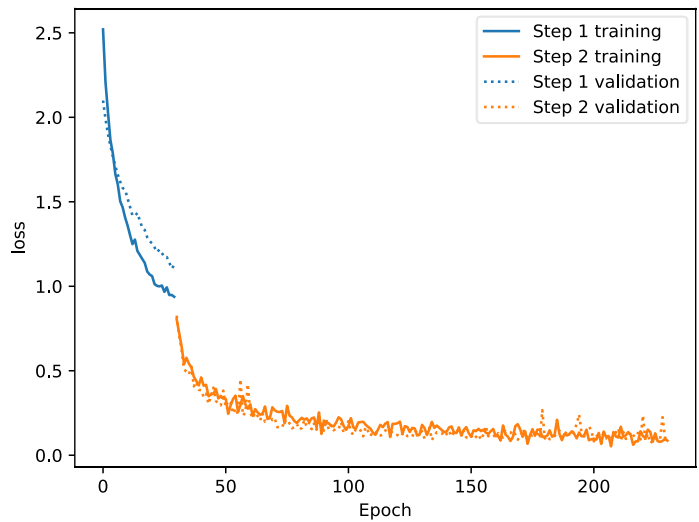

**Figure 9. Training and validation loss for Resnet-101 trained on UNIRI-SeaState-random-331.**

A summary of model sizes and parameters is shown in Table 3.

**Table 3.** Summary of Model Architectures.

| Deep Neural Network Architecture | Total parameters | Unfrozen layers | Total layers | Trainable parameters | Training epochs |
|---|---|---|---|---|---|
| ResNet-101 | 42,7 M | 305 | 345 | 24,8 M | 230 |
| ViT -b32 | 87,5 M | 14 | 19 | 21,3 M | 230 |

| MobileNetv2_1.00_224 | 2,7 M | 134 | 154 | 0,7 M | 430 |
| NASNet Mobile | 4,3 M | 649 | 769 | 1,6 M | 1.030 |

The input image size for all models was 224x224 pixels, while the images in the dataset were sized 331x331 pixels. This discrepancy was handled by further cropping the input image both during training and testing, however, during testing the crop position was always the same, from the center of the image, to ensure repeatability, while during training the crop position was random and was in effect a form of data augmentation.

During training, the following data augmentations were used:

- *Random cropping*: As noted above, random 224x224 crops were chosen from the initial 331x331 image in the dataset (Fig. 9). Note that depending on the dataset, the 331x331 images are themselves either random crops from the initial whole video frame or crops from the fixed bottom-left position. This augmentation was always applied.

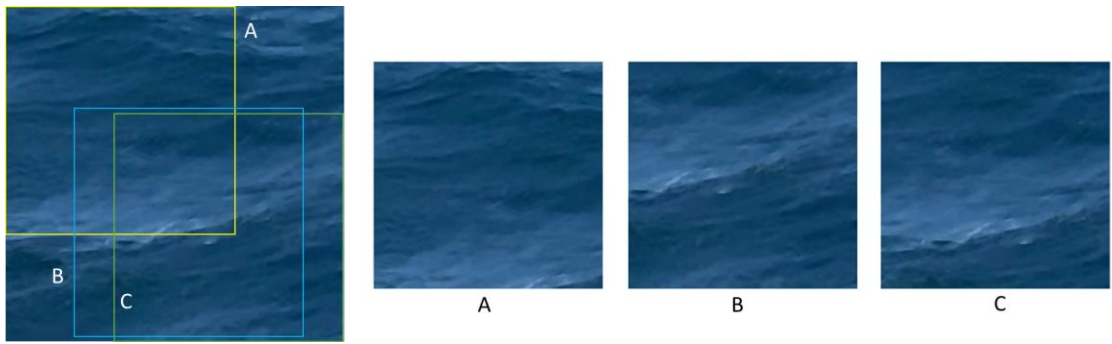

**Figure 10. Example of cropping data augmentation. A, B, and C show three random crop regions and corresponding cropped images.**

- *Motion blur*: A horizontal motion blur was applied to the batch with a 50% probability by convolving the image with a horizontal 7x7 motion blur kernel. This was chosen as a similar blur was often observed in the original images due to camera panning motion.

- *Horizontal flip*: images were flipped horizontally with 50% probability.

- *Random brightness and contrast adjustments*: Brightness was randomly scaled with a factor in the range [-0.2..2] and contrast was adjusted with a random factor in the range of [0.5..1.5]

- *Random rotation*: images were randomly rotated in a range of angles from -0.2pi to 0.2pi radians.

- *Random grayscale conversion:* Images were converted to grayscale with 20% probability.

In total 8 models were trained, one of each architecture for both datasets.

In order to test the generalization performance, the best-performing model was tested on the MU-SSiD (Umair et al. 2022) dataset, which contains images of sea states limited to the range of 1-4 Bft. Additionally, the ResNet model that performed best on our dataset was trained on the MU-SSiD dataset, with identical settings, and evaluated on both datasets.

### 2.5.    Evaluation Metrics

In the evaluation of our machine learning models, we have adhered to the standard metrics that are widely recognized in the field for assessing performance. These metrics are precision, recall, and the F1 score, each calculated for every individual class to capture the models' effectiveness at correctly identifying the distinct sea states.

Precision, also known as the Positive Predictive Value (PPV), is the proportion of true positive cases among all cases predicted as positive. It is a measure of a model's accuracy in identifying cases as belonging to a particular class. A higher precision indicates a lower rate of false positives. The formula for precision is given as:

$$Precision = \frac{TP}{TP + FP},$$
(1)

where TP represents the number of true positives, and FP denotes the number of false positives.

Recall (also known as sensitivity or True Positive Rate, TPR) assesses the model's ability to correctly identify all actual instances of a given class. It is particularly critical in scenarios where failing to detect a condition could have serious implications, and it addresses the question: Of all the actual sea states of a class, how many were identified?

$$Recall = \frac{TP}{TP + FN},$$
(2)

where FN is the number of false negatives. Both precision and recall are balanced by the F1 score, which is the harmonic mean of the two and is used when one needs to consider both false positives and false negatives equally.

The F1 serves as a balanced metric that harmonizes the precision and recall of a classification model into a single figure. This score is of paramount importance when one needs to consider the trade-off between the precision (or Positive Predictive Value, PPV) and the recall (True Positive Rate, TPR), such as in cases where both false positives and false negatives carry similar consequences. Moreover, the F1 Score is particularly informative in the presence of imbalanced class distributions, where the cost of misclassification can differ across classes. The F1 Score is calculated using the formula:

$$F1 = \frac{2\ x\ PPV\ x\ TPR}{PPV + TPR}.$$
(3)

In this formula, PPV is the number of true positive results divided by the number of all positive results predicted by the classifier, and TPR is the number of true positive results divided by the number of all actual positives. The F1 score effectively captures both the precision and robustness of the classifier's positive predictions.

For an overarching assessment of the model performance across all classes, we utilize Accuracy, which gives us the ratio of correctly predicted observations to the total observations. This is complemented by the Macro Average, which calculates metrics for each class independently and then takes the average, treating all classes equally.

$$\text{Macro Average} = \frac{1}{n} \sum_{\text{class}=1}^{n} \text{metric rate for each class}, \tag{4}$$

This means it does not take label imbalance into account. Conversely, the Weighted Average takes the label imbalance into account and calculates metrics for each class proportionally to its presence in the dataset.

$$\text{WA} = \frac{\sum_{\text{class}=1}^{n} \text{metric rate} * \ numer \ of \ instances \ per \ class}{total \ number \ of \ instances}, \tag{5}$$

By employing both macro and weighted averages, we ensure a comprehensive understanding of the model performance across classes with varying representations in the dataset. This dual approach allows us to recognize excellence in models' classification capabilities while also identifying potential areas of bias or weakness, particularly in classes that are underrepresented or more challenging to predict.

## 3. Results and Discussion

In our research, we conducted two distinct experiments to assess the efficacy of our selected machine-learning models in classifying sea states. The first experiment employed images sampled from random positions within the video footage, creating a dataset reflective of diverse perspectives. This dataset, referred to as "UNIRI-SeaState-R", is comprehensive, encompassing a total of 5,957 images, as previously detailed in Table 1. The second experiment, on the other hand, utilized images extracted from a fixed position in the video, specifically from the lower-left area relative to the ship's hull. This dataset, termed "UNIRI-SeaState-LL", consists of 5,970 images, ensuring consistency in the visual information presented to the models.

Both datasets were split into training and testing sets, with the training set comprising a significant majority (shown in Table 2). The testing set was curated to evaluate the models' performance objectively, as it contained previously unseen images that were representative of real-world conditions.

**Table 4. Performance Metrics for Models Trained and Tested on Images from Random Positions (UNIRI-SeaState-R Dataset)**

| Fine-tuned models | Metrics | Classes | | | | | | | | Accuracy | Macro avg | Weighted avg |
|---|---|---|---|---|---|---|---|---|---|---|---|---|
| | | 1 | 2 | 3 | 4 | 5 | 6 | 7 | 8 | | | |
| ResNet | precision | **1.000** | 0.904 | 0.538 | 0.923 | **1.000** | **0.802** | 0.963 | 0.878 | | **0.876** | **0.911** |
| | recall | 0.930 | 0.979 | **0.935** | **0.864** | 0.801 | 0.819 | 1.000 | 0.905 | **0.881** | **0.904** | **0.881** |
| | f1-score | 0.964 | **0.940** | 0.683 | **0.893** | 0.889 | **0.810** | 0.981 | 0.891 | | **0.881** | **0.888** |
| ViT | precision | 0.989 | 0.866 | **0.558** | 0.967 | 0.950 | 0.698 | 0.975 | **0.886** | | 0.861 | 0.895 |
| | recall | 0.935 | **0.983** | 0.886 | 0.707 | **0.866** | 0.867 | **0.996** | 0.979 | 0.868 | 0.902 | 0.868 |
| | f1-score | 0.961 | 0.921 | **0.685** | 0.817 | **0.906** | 0.773 | 0.985 | **0.930** | | 0.872 | 0.872 |
| MobileNet | precision | 0.970 | 0.417 | 0.337 | 0.544 | 0.589 | 0.275 | 0.727 | 0.341 | | 0.525 | 0.566 |
| | recall | 0.516 | 0.858 | 0.337 | 0.305 | 0.428 | 0.460 | 0.899 | 0.821 | 0.502 | 0.578 | 0.502 |
| | f1-score | 0.674 | 0.561 | 0.337 | 0.391 | 0.496 | 0.344 | 0.804 | 0.481 | | 0.511 | 0.501 |

| | | 1 | 2 | 3 | 4 | 5 | 6 | 7 | 8 | Accuracy | Macro avg | Weighted avg |
|---|---|---|---|---|---|---|---|---|---|---|---|---|
| NASNet | precision | 0.960 | **0.910** | 0.494 | 0.887 | 0.843 | 0.472 | **0.991** | 0.789 | | 0.793 | 0.823 |
| | recall | **0.976** | 0.907 | 0.776 | 0.563 | 0.733 | 0.794 | 0.983 | 0.947 | 0.781 | 0.835 | 0.781 |
| | f1-score | **0.968** | 0.908 | 0.603 | 0.689 | 0.784 | 0.592 | **0.987** | 0.861 | | 0.799 | 0.787 |

Upon reviewing the performance metrics presented in Table 4, it is evident that ResNet achieves the highest overall accuracy, closely followed by Vision Transformer (ViT). This high accuracy demonstrates the models' ability to correctly classify sea states across the dataset. Notably, the performance of NASNet and MobileNet is marginally lower, which could be attributed to their significantly smaller number of parameters—approximately ten times less than ResNet—manifesting in a roughly 10% drop in accuracy for NASNet.

All models exhibit the highest F1 scores for class 7, despite it not being the most represented class in the dataset. This suggests that the images for class 7 are of sufficient quality to facilitate effective learning by the models. Conversely, all models perform the least effectively in class 3, followed by class 6, indicating these classes may benefit from a more diversified or expanded set of training examples to enhance model learning.

The class-wise variation in precision and recall further illustrates the performance of each model. Class 7, with its high precision across all models, suggests that the class contains distinctive, well-captured features conducive to accurate classification. High recall scores in other classes indicate successful identification of the most positive examples within those classes.

In sum, ResNet's superior accuracy, alongside its strong macro and weighted average scores, showcases its robustness and adaptability across varied sea states. In comparison, the smaller architectures of both NASNet and MobileNet still deliver respectable average precision and recall, but with significantly lower scores for some classes.

On the set of images sampled from a fixed position, all models show better results than in the case of random sampling (Table 5). The best results are again achieved by ResNet and ViT with 97% and 96% average F1-score. The models achieve uniform F1 results for each of the classes, namely ResNet in the range from 93 to 100%, and ViT in a slightly wider range from 86% to 99%. Both models achieve strong macro-average scores and show excellent results for all classes concerning all metrics except for two, classes 3 and 6 suggesting that their performance is evenly distributed across classes.

**Table 5. Evaluation of Models Trained and Tested on Images Sampled from a Fixed Lower-Left Position (UNIRI-SeaState-LL Dataset)**

| Fine-tuned models | Metrics | Classes | | | | | | | | Accuracy | Macro avg | Weighted avg |
|---|---|---|---|---|---|---|---|---|---|---|---|---|
| | | 1 | 2 | 3 | 4 | 5 | 6 | 7 | 8 | | | |
| ResNet | precision | **1.000** | 0.941 | **0.911** | 0.981 | **0.996** | 0.952 | **1.000** | 0.944 | **0.974** | 0.966 | **0.975** |
| | recall | 0.964 | **1.000** | **0.959** | 0.962 | 0.986 | 0.936 | **1.000** | **1.000** | **0.974** | 0.976 | **0.974** |
| | f1-score | 0.981 | 0.970 | **0.934** | 0.971 | 0.991 | 0.944 | **1.000** | 0.971 | **0.974** | 0.970 | **0.974** |
| ViT | precision | 0.997 | **0.966** | 0.777 | **0.991** | 0.989 | 0.899 | 0.983 | 0.943 | 0.955 | 0.943 | 0.959 |
| | recall | **0.990** | 0.997 | 0.955 | 0.899 | 0.964 | 0.911 | **1.000** | 0.988 | 0.955 | 0.963 | 0.955 |
| | f1-score | **0.993** | **0.981** | 0.857 | 0.942 | 0.976 | 0.905 | 0.991 | 0.965 | 0.955 | 0.951 | 0.956 |
| MobileNet | precision | 0.991 | 0.789 | 0.615 | 0.921 | 0.979 | 0.508 | 0.972 | 0.851 | 0.844 | 0.828 | 0.870 |
| | recall | 0.875 | 0.934 | 0.861 | 0.722 | 0.898 | 0.647 | **1.000** | 0.952 | 0.844 | 0.861 | 0.844 |
| | f1-score | 0.929 | 0.855 | 0.718 | 0.810 | 0.937 | 0.569 | 0.986 | 0.899 | 0.844 | 0.838 | 0.850 |
| NASNet | precision | **1.000** | 0.879 | 0.814 | 0.934 | 0.976 | 0.626 | 0.988 | 0.932 | 0.902 | 0.894 | 0.915 |
| | recall | 0.917 | 0.983 | 0.788 | 0.856 | 0.916 | 0.877 | 0.994 | 0.976 | 0.902 | 0.913 | 0.902 |

| f1-score | 0.957 | 0.928 | 0.801 | 0.893 | 0.945 | 0.730 | 0.991 | 0.953 | 0.902 | 0.900 | 0.905 |

The other models for the same classes also show the worst performance. But, although NASNet has 10 and 20 times fewer parameters than the ResNet and ViT networks, respectively, for 6 out of 8 classes they achieve an F1 score above 90%, while for classes 3 and 6, for which the mentioned models have somewhat lower accuracy, they achieve an F1 score of 80% and 73%, which is a respectable result considering the complexity of the task.

MobileNet is significantly the smallest network, with the smallest number of parameters and with twice as many iterations as the best networks it manages to achieve satisfactory results with an 85% average F1 score which shows that it can be used successfully on edge devices with limited computing resources.

Analysis of performance metrics in Table 5 reveals that all models, despite different architectures and a number of parameters, show potential for high performance in sea state classification, taking into account not only overall accuracy but the impact of class distribution on these measures and performance for each class.

## 3.1. Evaluation of the ResNet model on a different dataset of sea state images

The MU-SSiD dataset (Umair, et al. 2022) is a publicly available dataset of sea state images in four classes from 1-4 Bft. The dataset contains 6000 images per class cropped from a source video in full HD resolution, similar to our dataset.

We trained the ResNet model that performed best on our dataset on the MU-SSiD dataset using the same training setup as when training on the UNIRI-SeaState-LL dataset, only with the adjusted number of classes in the final layer, 4 in this case. In the following, we refer to this model as ResNet-Mu-SSiD.

We compared the performance of the models trained on one dataset to the other dataset to determine the robustness of the models on different camera settings and applicability on sea state images collected in very different conditions. Namely, we test the ResNet model trained on the UNIRI-SeaState-LL dataset (ResNet-UNIRI) on the UNIRI-SeaState-LL and MU-SSiD datasets, and similarly, we test the ResNet-Mu-SSiD on both the UNIRI-SeaState-LL and MU-SSiD datasets.

The results for the ResNet-Mu-SSiD model are shown in Table 6. The model performs well when tested on the images from a dataset similar to the one it was trained on, and similarly to the reported results (Umair, et al. 2022). However, in the UNIRI-SeaState-LL dataset, the model ResNet-Mu-SSiD shows a significant drop in performance, from 78% accuracy to 31% accuracy.

**Table 6. Evaluation of ResNet Model trained on MU-SSiD training dataset and tested on both test set of the MU-SSiD and test set of the UNIRI-SeaState-LL**

| Test set | Metrics | Classes | | | | Accuracy | Weighted avg | Performance drop |
|---|---|---|---|---|---|---|---|---|
| | | 1 | 2 | 3 | 4 | | | |
| MU-SSiD | precision | 0,93 | 0,59 | 0,67 | 0,99 | | 0,7928 | - |
| | recall | 0,77 | 0,68 | 0,67 | 1 | 0,78 | 0,7794 | - |
| | f1-score | 0,84 | 0,63 | 0,67 | 0,99 | | 0,7833 | - |
| UNIRI-SeaState-LL | precision | 0,21 | 0,43 | 0,67 | 0,63 | | 0,491 | 0,3 |
| | recall | 0,48 | 0,85 | 0,12 | 0,02 | 0,31 | 0,3133 | 0,47 |
| | f1-score | 0,29 | 0,57 | 0,21 | 0,04 | | 0,2339 | 0,55 |

The results for the ResNet-UNIRI model are shown in Table 7. Similarly, as the ResNet-Mu-SSiD model, the model performs well when tested on the test subset of the dataset it was trained on. Again, the performance drops significantly when tested on different datasets (MU-SSiD), down to 28% accuracy from 88% accuracy. It can be noted that this model was trained on 8 classes, and thus could classify an image from the test set into classes that were not present in the dataset at all (5-8 Bft).

**Table 7. Evaluation of ResNet Model trained on UNIRI-SeaState-LL training dataset and tested on both test set of the MU-SSiD and test set of the UNIRI-SeaState-LL**

| Test set | Metrics | Classes | | | | | | | | Accuracy | Weighted average | Perform. drop |
|---|---|---|---|---|---|---|---|---|---|---|---|---|
| | | 1 | 2 | 3 | 4 | 5 | 6 | 7 | 8 | | | |
| UNIRI-SeaState-LL | precision | 1 | 0,9 | 0,54 | 0,92 | 1,00 | 0,80 | 0,96 | 0,88 | | 0,91 | |
| | recall | 0,93 | 0,98 | 0,93 | 0,86 | 0,80 | 0,82 | 1,00 | 0,91 | | 0,88 | |
| | f1-score | 0,96 | 0,94 | 0,68 | 0,89 | 0,89 | 0,81 | 0,98 | 0,89 | 0,88 | 0,89 | |
| MU-SSiD | precision | 1 | 0,4 | 0,78 | 0 | 0 | 0 | 0 | 0 | | 0,55 | 0,36 |
| | recall | 0,05 | 0,91 | 0,14 | 0 | 0 | 0 | 0 | 0 | | 0,28 | 0,61 |
| | f1-score | 0,09 | 0,56 | 0,24 | 0 | 0 | 0 | 0 | 0 | 0,28 | 0,22 | 0,67 |

The confusion matrix for ResNet-UNIRI tested on the MU-SSiD dataset (Table 8), shows the problem of misclassification. For example, many instances of class 4 in the MU-SSiD dataset were classified as class 7. Also, many instances of class 3 were classified as class 2. The reason for both errors might be explained by the difference in data collection, primarily the camera position and height. Since the images in the UNIRI-SeaState-LL dataset were taken from a much greater height of about 40 m, the difference between calmer sea states e.g. 2 and 3 are not as pronounced as when captured from 2-3 m.

**Table 8. Confusion matrix of ResNet Model trained on UNIRI-SeaState-LL training dataset and tested on the MU-SSiD test set**

| True class | Predicted class | | | | | | | |
|---|---|---|---|---|---|---|---|---|
| | 1 | 2 | 3 | 4 | 5 | 6 | 7 | 8 |
| 1 | 56 | 655 | 47 | 1 | 35 | 238 | 4 | 164 |
| 2 | 0 | 1091 | 2 | 0 | 19 | 15 | 0 | 73 |
| 3 | 0 | 945 | 173 | 12 | 0 | 69 | 0 | 1 |
| 4 | 0 | 13 | 0 | 0 | 0 | 0 | 1187 | 0 |
| 5 | 0 | 0 | 0 | 0 | 0 | 0 | 0 | 0 |
| 6 | 0 | 0 | 0 | 0 | 0 | 0 | 0 | 0 |
| 7 | 0 | 0 | 0 | 0 | 0 | 0 | 0 | 0 |
| 8 | 0 | 0 | 0 | 0 | 0 | 0 | 0 | 0 |

The authors (MU-SSiD) state that the raw videos for their data set were recorded at a height of 2 to at most 12 m, while our data set was formed from videos taken at a height of about 40 m. This means that the images in these datasets were collected at a significantly different setup (camera position), which varies in height from 4 to 20 times. Thus, the waves of e.g. sea state 4 appear visually much smaller on our dataset, and waves of sea state 7 in our dataset visually appear similar to sea state 4 taken from a much lower height as in the MU-SSiD dataset (Fig. 11).

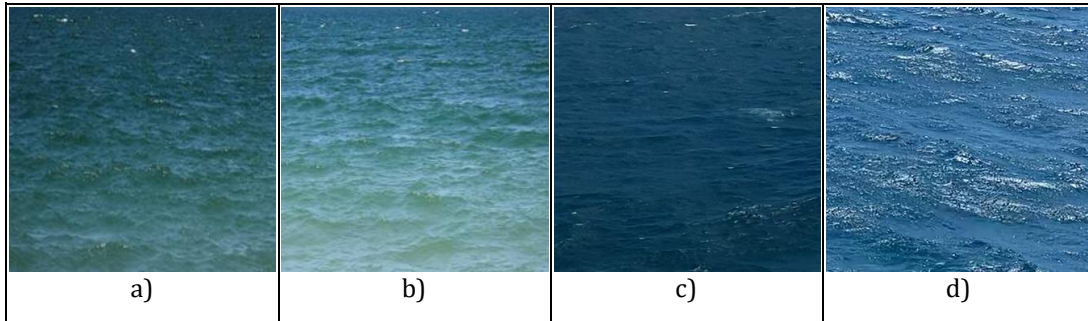

Figure 11. a), b) samples of sea state 4 in MU-SSiD dataset; c), d) samples of sea state 7 in UNIRI-SeaState-lower-left-331 dataset.

The results suggest that model accuracy is influenced by recording conditions, including camera height, grazing angle, and the size of the individual crops, pointing to an area for further research. The observed discrepancies in wave appearance between datasets captured at different heights underscore the need for model adaptation to account for varying ship sizes and other related factors.

## 3.2. Evaluation of the ResNet model with various training set sizes and input image size

To determine the sufficient number of training images per class in the UNIRI-SeaState-R dataset, a preliminary experiment was done using a varying number of input images to train the model with ResNet-101 architecture. Here, the number of training images per class was increased from 10 to 750 in several steps, roughly doubling the number of images in each step.

The experiment was repeated with an input image size of 331x331 without further cropping. Besides the training size, the training and testing regime was identical to the experiments above. The performance of the models is shown in Figure 12., and the corresponding training times in Figure 13.

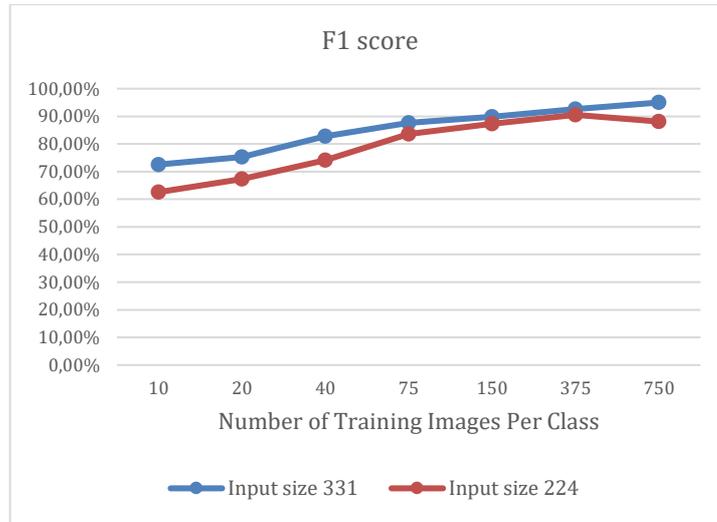

**Figure 12. F1 scores for the ResNet model trained on subsets of the UNIRI-SeaState-R dataset with different number of images per class**

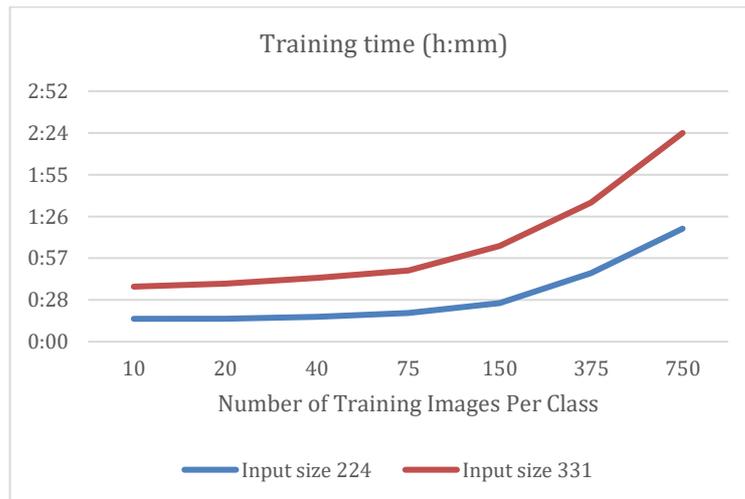

**Figure 13. Training times for ResNet models trained on subsets of the UNIRI-SeaState-R dataset.**

It can be seen that the classification performance for the input size of 224x224 no longer increases after 375 images. Since the performance still increased with 750 images per class for a 331x331 input size, this number was chosen for the final dataset. Simultaneously, training time increases significantly with doubling the number of training images and with increasing the image size (Fig. 13).

### 3.3. Comparison of models training and inference requirements

Besides the performance on new data, resource requirements, especially training time and memory usage may be of concern when choosing a model. Table 9 presents the times spent on training the models on the UNIRI-SeaState-LL training dataset for different model architectures. The training was performed on a PC with Intel Core i9-10900 10-core CPU with 64 GB of RAM and a single NVIDIA 3090 GPU with 24 GB of RAM. The dataset was stored on a Crucial MX500 2TB SATA SSD with a declared 560MB/s maximum read speed. The batch size was chosen manually in a few attempts so that the training and pre-processing could be performed on the GPU without exhausting the GPU memory. The software environment consists of the Ubuntu 20.04.6 LTS operating system, NVIDIA CUDA software version 12.2 and driver version 535.171, Python 3.10.8, and Tensorflow 2.11.0.

**Table 9. Training parameters and times for training on the UNIRI-SeaState-LL training dataset**

| Model | Input Image Size | Training Batch Size | Epochs | Total training time (h: mm) | Training Time Per Epoch (s) |
|---|---|---|---|---|---|
| ResNet | | 250 | 230 | 1:18 | 20 |
| ViT | 224x224 | 200 | 230 | 1:19 | 21 |
| MobileNet | | 250 | 430 | 1:48 | 15 |
| NASNet | | 250 | 1030 | 4:24 | 15 |
| ResNet | 331x331 | 20 | 230 | 2:24 | 38 |

It can be noted that the models with a similar number of parameters took similar time per epoch to train, for the same input image size. However, the models with fewer parameters needed a larger number of epochs to train, so the more complex models that performed best, ResNet and ViT, took the least total time to train. The simpler MobileNet model took somewhat longer to train, yet didn't reach a similar performance level. The NASNet model, however, reached very good performance despite the much smaller number of parameters than either ResNet or ViT but took significantly longer to train, more than three times as much. Training with input image size increased from 224x224 to 331x331 almost doubling the training size for ResNet.

Resources used during inference and average inference times are shown in Table 10. Memory usage for the model was determined using TensorFlow's *get_memory_info()* function after relevant steps (loading the model, after inference).

**Table 10. Inference characteristics of different classification models.**

| Model | Input Image Size | Inference Batch Size | GPU Memory usage (model only, MB) | Peak GPU Memory usage during inference (MB) | Average inference throughput (images/s) |
|---|---|---|---|---|---|
| ResNet | | 250 | 243 | 4,087.95 | 376.41 |
| ViT | 224x224 | 250 | 420.7 | 877.82 | 476.36 |
| MobileNet | | 250 | **9.9** | **557.07** | **634.34** |
| NASNet | | 250 | 24.4 | 1,867.89 | 451.88 |
| ResNet | 331x331 | 250 | 243 | 12,225.22 | 226.30 |

The best-performing model, ResNet, is also the slowest and most memory-demanding during inference. Notably, the ViT model with a similar number of parameters and performance in terms of F1 score performs faster and with much smaller peak memory usage during inference, making it a very attractive alternative if resources are constrained. The least resource-demanding and fastest model is MobileNet, however with significantly lower classification performance. Considering that speed of inference is likely not critical for sea state detection where classification is not expected to be continuous and real-time, the use of this model seems justified only in the most memory-constrained situations.

## 3.4. Discussion

The comparative analyses of model performance based on images sampled from fixed and random positions have revealed several key considerations for the classification of sea states.

The empirical results show that the models trained on the images from a fixed lower-left position achieve better results than those trained on randomly sampled images. This may be attributed to the camera angle and the distance from the camera, which can make waves farther away appear similar to closer waves from a different class. Such observations underscore the model limitation due to dependence on the camera's height, distance, and orientation. This dependency is further confirmed when training and testing on datasets captured from different heights; models trained at one height struggle to successfully recognize the sea state captured from substantially different heights. Subsequent studies should investigate the feasibility of a generally acceptable model for sea state classification or confirm the need for distinct models trained to specific camera heights.

It emerges that while the inherent complexity of a model, inferred from the number of its parameters, contributes to the granularity of data interpretation, the strategy for image sampling also significantly influences the performance metrics. A deliberate approach to the collection of training data, encompasses not only a comprehensive spectrum of sea states but also other factors such as time of day to improve the robustness and universality of these models.

There is also an opportunity to improve the model accuracy, especially for sparsely represented sea states, by targeted augmentation of the dataset with additional instances from these less prevalent conditions.

Looking ahead, we aim to extend the dataset to capture a wider variety of sea states and to continuously refine our model. This progression will allow us to further solidify the models' performance, ensuring accurate sea state recognition that is vital for safe and efficient maritime operations.

## 4. Conclusion

This research explores the application of computer vision and deep learning to the automated real-time classification of sea states, contributing to the field of maritime safety and vessel energy efficiency. Advanced models, including transfer learning, have effectively categorized sea states within the operational envelope based on the Beaufort scale.

As part of the research, relevant video data of the state of the sea were collected and two sets of UNIRI-SeaState-LL and UNIRI-SeaState-R of data were formed, sampled from the video according to different strategies, and prepared for supervised machine learning of the model. The sets cover eight classes of sea state within an operational framework based on the Beaufort scale, each marked from 1 to 8.

We selected state-of-the-art models of deep neural networks of different architectures and parameter ranges from 2.7 to 87 M, MobileNet, NASNet, ResNet, and ViT, and fine-tuned them using transfer learning and a custom data set to recognize different sea states.

To ascertain the efficacy of selected models for sea state classification, we evaluated and contrasted their performance using standard metrics: precision, recall, F1 score, and accuracy, considering both macro and weighted average results.

Our research demonstrates that the ResNet model delivered the optimal performance across all metrics, with macro and weighted average results considered, achieving an average F1 score of 97%. The ViT model, which has twice as many parameters, achieved comparative results with an average F1 score of 95.6%. Two other networks with a significantly smaller number of parameters, ranging from 10-40 times less, also achieved good results and attained an average F1-score of 90% for NASNet while MobileNet reached 85%.

The study found that model performance is influenced by image capture strategy, with models trained on images from a stable position performing better than those with randomly sampled images. Additionally, variations in image capture height were shown to affect model accuracy, due to the altered appearance of sea states at different recording points. These findings have significant implications for growing research in autonomous navigation, emphasizing the need for real-time, accurate sea state assessment to ensure the safety and efficiency of unmanned maritime vessels in the push toward autonomous maritime operations.

In future work, we will focus on broadening the scope of data collection to enhance the representativeness of the dataset, thereby reinforcing the model's estimation performance and generality. Furthermore, we intend to treat the sea state recognition problem as a regression problem or as a fuzzy class to predict sea states as they are usually done in practice. Additionally, we aim to explore the development of a model for sea state classification that either accounts for or is invariant to recording conditions. Such a model would perform well across various recording positions and other geometrical aspects, reducing the need for ship-specific fine-tuning.

## Acknowledgments

This research was partially supported by the HORIZON EUROPE Widening INNO2MARE project (grant agreement ID: 101087348).